\documentclass[10pt,twocolumn,letterpaper]{article}

\usepackage{iccv}              

\definecolor{iccvblue}{rgb}{0.21,0.49,0.74}
\usepackage[pagebackref,breaklinks,colorlinks,allcolors=iccvblue]{hyperref}

\usepackage[table]{xcolor}
\usepackage{appendix}
\usepackage{breqn}
\usepackage{csquotes}
\usepackage{xspace}
\usepackage{mathtools}
\usepackage{graphicx}
\usepackage{booktabs}
\usepackage{multirow}
\usepackage{adjustbox}
\usepackage{tabularx}
\usepackage{tablefootnote}
\usepackage{float}
\usepackage{colortbl}
\usepackage{arydshln}
\usepackage{amssymb}

\newcommand*{\lora}{\textsc{LoRA}\xspace}
\newcommand*{\vpt}{\textsc{VPT}\xspace}
\newcommand*{\bias}{\textsc{BitFit}\xspace}
\newcommand*{\bitfit}{\textsc{BitFit}\xspace}
\newcommand*{\full}{\textsc{Full}\xspace}

\newcommand*{\apla}{\textsc{APLA}\xspace}

\newcommand*{\dinovtwo}{\textsc{DinoV2}\xspace}

\newcommand*{\ssa}{\textsc{SSA}\xspace}
\newcommand*{\sa}{\textsc{SA}\xspace}
\newcommand*{\essa}{\textsc{ESSA}\xspace}
\newcommand*{\peft}{\textsc{PEFT}\xspace}
\newcommand*{\knn}{$k$-NN\xspace}
\newcommand*{\imagenettwentyonek}{\textsc{ImageNet}-21K\xspace}

\newcommand*{\vit}{\textsc{V}i\textsc{T}\xspace}

\newcommand*{\vitl}{\textsc{V}i\textsc{T-L}\xspace}

\newcommand*{\NumDatasets}{5\xspace}
\newcommand*{\MemReduction}{40.1\%\xspace}
\newcommand*{\ThroughputIncrease}{25.2\%\xspace}

\definecolor{pinkcolor}{RGB}{255, 105, 180}  

\def\paperID{17} 
\def\confName{ICCV}
\def\confYear{2025}

\title{
Efficient Self-Supervised Adaptation for Medical Image Analysis
}

\author{\noindent
{Moein Sorkhei} $^{1,2}$ 
\hspace{-1.5mm}\textsuperscript{
\thanks{Corresponding author: Moein Sorkhei \textless{}sorkhei@kth.se\textgreater{}}
}
{Emir Konuk} $^{1,2}$
{Jingyu Guo} $^{1,2}$
{Chanjuan Meng} $^{1,2}$
{Christos Matsoukas} $^{1,2}$
\\
{Kevin Smith} $^{1,2}$ \\[1mm]
\normalfont{\noindent
$^{1}$ KTH Royal Institute of Technology, Stockholm, Sweden} \\
$^{2}$ Science for Life Laboratory, Stockholm, Sweden 
}

\begin{document}
\maketitle
%
\begin{abstract}

Self-supervised adaptation (\ssa) improves foundation model transfer to medical domains but is computationally prohibitive. Although parameter efficient fine-tuning (\peft) methods such as \lora have been explored for supervised adaptation, their effectiveness for SSA remains unknown. 
In this work, we introduce efficient self-supervised adaptation (\essa), a framework that applies parameter-efficient fine-tuning techniques to SSA with the aim of reducing computational cost and improving adaptation performance. 

To the best of our knowledge, we are the first to demonstrate that \peft methods can be effectively applied to \ssa to improve self-supervised learning, challenging the  assumption that full-parameter \ssa is necessary for optimal performance. Furthermore, we show that applying \peft during supervised adaptation following self-supervision leads to additional performance gains, outperforming full-parameter training.

\end{abstract}

\section{Introduction}

\begin{figure*}[t]
    \centering
    \begin{tabular}{cc}
    \multicolumn{2}{c}{\includegraphics[width=1.0\linewidth]{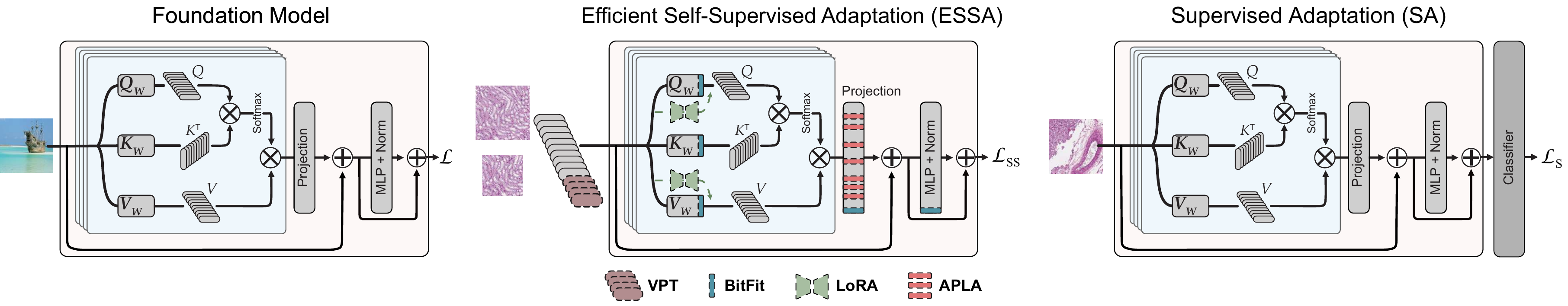}}
    \\
    \multicolumn{2}{c}{\scriptsize{
    (a) \hspace{40mm} (b) \hspace{40mm} (c)
    }}\\[3mm]
    \includegraphics[width=0.48\linewidth]{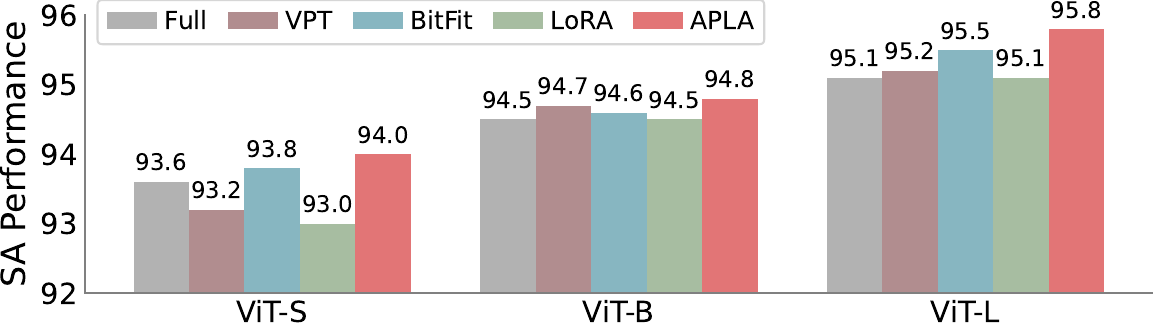} 
    &
    \includegraphics[width=0.48\linewidth]{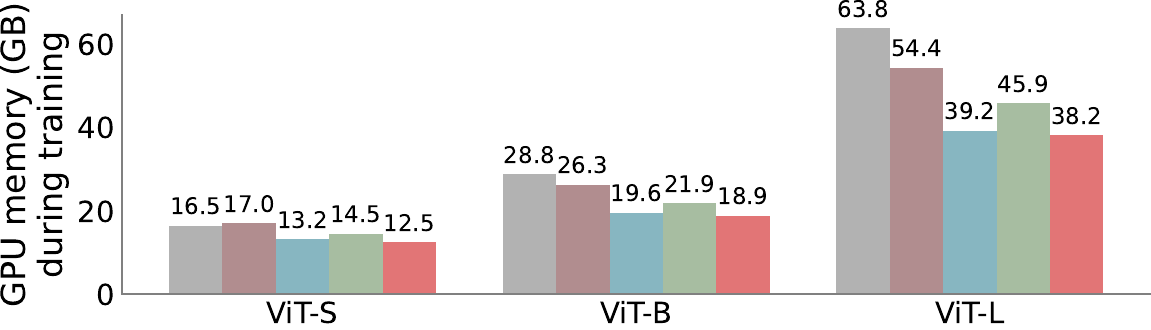}
    \\[-1mm]
    \scriptsize{(d)} &
    \scriptsize{(e)}\\[-1mm]
    \end{tabular}
    \caption{
    \textit{(Top) The workflow for \essa.} (a) Starting with a foundation model, (b) we investigate different \peft methods for efficient self-supervised adaptation (\essa) to adapt the model to the target domain, (c) followed by full-parameter supervised fine-tuning to the target task (\sa). 
    \textit{(Bottom)} (d) performance and (e) GPU memory consumption of \peft methods during \essa compared to full-parameter \ssa.   
    }
    \label{fig:fig1}
\end{figure*}

Foundation models pretrained on diverse, large-scale natural scene datasets have shown remarkable effectiveness in medical imaging \cite{bommasani2021opportunities,huix2024natural}. 
However, their direct transfer is hindered by domain shifts, as natural and medical images differ significantly in structure and content \cite{azizi2021big,raghu2019transfusion,matsoukas2022makes}. 
Transfer learning, or supervised adaptation (\sa), helps bridge this gap by fine-tuning the model weights to the new domain. 
Further gains can be achieved by first applying self-supervised adaptation (\ssa), where the model is pretrained on unlabeled images from the target domain before fine-tuning using supervision  \cite{azizi2021big,kumari2024deep,sowrirajan2021moco,matsoukas2023pretrained,xiao2023delving} (Figure \ref{fig:fig1}). 
A drawback of SSA is its high computational demand, requiring significant memory and long training times \cite{azizi2021big}, which makes it impractical for large foundation models.

Parameter efficient fine-tuning (\peft) methods such as \lora \cite{hu2021lora} and \vpt \cite{jia2022visual} have been developed to reduce the costs of \sa, but it remains to be seen whether these methods can effectively optimize \ssa objectives without degrading the representation quality.
In this paper, we introduce the setting of \textit{efficient self-supervised adaptation} (\essa), where \peft techniques are applied to \ssa with the aims of reducing computational cost and enhancing downstream performance. 
We evaluate several state-of-the-art methods for \essa over a variety of medical imaging tasks. 
We find that for sufficiently large foundation models, \essa  outperforms full-parameter \ssa, challenging assumptions behind the current practice of full-network training.
\essa can reduce GPU memory consumption by up to \MemReduction and increase training throughput by \ThroughputIncrease, without affecting inference costs.

We also investigate the effectiveness of \peft for test-time training (TTT), where the model is further adapted to distribution shifts at inference time using self-supervision. 
We find that \peft methods can be successfully applied in this setting to improve adaptation during inference while reducing computational cost.

Our contributions are as follows:
\begin{itemize}
    \item We systematically evaluate \peft methods applied to \ssa, demonstrating that they improve performance over full-parameter \ssa for sufficiently large foundation models.
    \item We show that applying \peft during \sa following \ssa leads to additional performance gains, outperforming full-parameter training.
    \item We show that \essa methods outperform full-parameter tuning in test-time training (TTT), making them particularly effective in low-label settings.
\end{itemize}

The code to reproduce our experiments can be found at \href{https://github.com/MoeinSorkhei/APLA}{
\texttt{\textcolor{pinkcolor}{github.com/MoeinSorkhei/APLA}}}.

\section{Related Work}

Supervised adaptation (\sa), or transfer learning from models pretrained on natural image datasets, is widely used in medical imaging due to the scarcity of labeled data \cite{bommasani2021opportunities,huix2024natural}.
However, this approach faces two major challenges:
(1) a domain gap between natural and medical images, which limits transfer learning performance \cite{azizi2021big,matsoukas2022makes,raghu2019transfusion}, and
(2) the growing computational burden as foundation models \cite{bommasani2021opportunities} become more widely adopted and continue to scale in size \cite{oquab2023dinov2,dehghani2023scaling}.

\ssa offers improved performance by self-supervised training on unlabeled images from the target domain before supervised adaptation \cite{azizi2021big,kumari2024deep,sowrirajan2021moco,matsoukas2023pretrained,xiao2023delving}, with the added benefit of avoiding the problem of label scarcity.
\ssa methods in the past have relied on task-specific pretext tasks \cite{cui2024endodac}, adversarial learning \cite{xu2019self}, or feature alignment \cite{zhang2019whole}
but more recently, task-agnostic self-supervised approaches, 
such as contrastive \cite{azizi2021big} and discriminative \cite{matsoukas2021time} learning,
have demonstrated strong performance and broad applicability for \ssa with foundation models.
Despite their success, these methods significantly increase computational demands when adapting large-scale models \cite{azizi2021big}.

To mitigate the computational costs of full fine-tuning, \peft methods have been widely explored in \sa.
Techniques such as \lora \cite{hu2021lora}, \vpt \cite{jia2022visual}, \bitfit \cite{zaken2021bitfit}, and \apla \cite{sorkhei2025apla}
have successfully lowered resource requirements by limiting the number of trainable parameters.
While these methods effectively reduce training costs for supervised fine-tuning, they are not guaranteed to be effective for \ssa, as the objective is to learn reusable, domain-general representations rather than task-specific ones, which means that adapting fewer parameters could be risky, \ie it might harm general feature learning.
There have been attempts to apply \peft to standard transfer learning in the medical domain \cite{zhu2024melo, dutt2023parameter, lian2024less}.
However, to the best of our knowledge, the use of \peft strategies for \ssa has not been systematically investigated until now.

\section{Methods}

We propose efficient self-supervised adaptation (\essa), a parameter-efficient adaptation strategy for foundation models in medical imaging. 
\essa applies PEFT in the context of a two-stage foundation model adaptation framework.
This starts from a pretrained foundation model depicted in Figure \ref{fig:fig1}a:

\begin{enumerate}
\item \textbf{Self-Supervised Adaptation (SSA)}: The foundation model is first adapted to the target domain using task-agnostic self-supervised learning on unlabeled images (Fig.~\ref{fig:fig1}b).
\item \textbf{Supervised Adaptation (SA)} The domain-adapted model is then fine-tuned with an added predictor on labeled data for downstream tasks (Fig.~\ref{fig:fig1}c).
\end{enumerate}
We extend the benefits of {\peft}s beyond this two-stage framework by exploring their use in test-time training (TTT) \cite{sun2020test} as an optional step that allows dynamic adaptation to unlabeled out-of-domain samples during inference.

\subsection{Two-stage \ssa and \sa adaptation framework}
\emph{We begin by describing the two-stage adaptation framework shown in Figure~\ref{fig:fig1} without incorporating parameter-efficient fine-tuning (\peft).}
The first stage (Fig.~\ref{fig:fig1}b) aligns the foundation model’s representations with the medical domain using self-supervised learning. 
This enables the model to learn rich representations from the target domain, but it is task-agnostic, limiting its ability to support diagnostic tasks. 

This necessitates a second stage (Fig.~\ref{fig:fig1}c), supervised adaptation (\sa), where a classifier or task-specific head is added to the domain-adapted feature extractor from stage one. 
The model is then fine-tuned using supervised learning on image-label pairs to optimize task performance. 

\subsection{Parameter-efficient fine-tuning (\peft)}
Full fine-tuning of large foundation models is computationally expensive. 
To mitigate this cost, \peft methods update only a small subset of parameters while keeping most of the model frozen. 
These approaches fall into two broad categories: additive methods \cite{hu2021lora,jia2022visual}, which introduce additional trainable components without modifying the backbone, and parameter-selective methods \cite{zaken2021bitfit,sorkhei2025apla} which fine-tune a limited subset of existing model weights. 
\textit{While \peft has been successfully applied for standard transfer learning, its effectiveness for \ssa remains unexplored.}

\subsection{Efficient self-supervised adaptation (\essa)}
We apply \peft techniques to \ssa (and, optionally \sa) optimizing only a small set of parameters, $\gamma$, instead of the full foundation model parameters. 
The general form of the efficient \ssa step (\essa) is given by
\begin{equation}
    \min_{\gamma} \mathcal{L_{SS}}\Bigl(F_{\phi^*,\gamma}(x)\Bigr)
    \label{eq:eff_ssa}
\end{equation}
which yields a model \(\hat{F}\) adapted to the target domain.
This is followed by an \sa step, which can optionally also be made efficient with {\peft}s by
\begin{equation}
    \min_{\theta,\gamma} \mathcal{L_{S}}\Bigl(y,h_\theta(\hat{F}_{\phi^*,\gamma}(x))\Bigr)
    \label{eq:eff_sa}
\end{equation}
where \(h\) denotes a prediction head with parameters \(\theta\).
We use \(\mathcal{L_{SS}}\) to indicate a self-supervised objective calculated on images \( x\sim \mathcal{D}\) from the target dataset,
and \(\mathcal{L_{S}}\) is a supervised loss function calculated on image-label pairs \( (x,y)\sim \mathcal{D}\) from the same dataset\footnote{Note that frozen \(\phi^*\) and tunable parameters \(\gamma\) of the foundation model \(F\) and domain adapted model \(\hat{F}\) can be different, \eg a different adapter block or a different subset of existing parameters may be tuned during the \ssa and \sa stages.}.
For some efficient adaptation methods like \lora \cite{hu2021lora} and \vpt \cite{jia2022visual},
all parameters of the foundation model \(F_\phi\) are frozen,
\(\phi^* = \phi\),
and the tunable parameters \(\gamma\) represent the additional parameters injected to the model.
For \lora \cite{hu2021lora}, these are low-rank matrix approximations to the changes in model parameters during adaptation, 
whereas for \vpt \cite{jia2022visual} they are a learnable prompt vector prepended to the foundation model.
For other efficient methods like \bitfit \cite{zaken2021bitfit} and \apla \cite{sorkhei2025apla}, 
tunable \(\gamma\) parameters represent a subset of the foundation model parameters, \(\phi = \phi^* \cup \gamma \),
while the rest, \(\phi^*\), are kept frozen.
For \bitfit \cite{zaken2021bitfit}, the tuneable subset \(\gamma\) are the bias terms 
while \apla \cite{sorkhei2025apla} selectively fine-tunes a random subset of the parameters in the layer immediately following the attention mechanism, which its authors identified as critical for adaptation.

\subsection{Test-time training (TTT)}
Test-time training is a self-supervised domain adaptation method \cite{sun2020test,gandelsman2022test} that updates the feature extractor during inference using self-supervised learning, while keeping the classifier fixed. Instead of full-network adaptation, we apply \peft methods to optimize a small subset of parameters of the feature extractor. 
This enables efficient on-the-fly adaptation to domain shifts without explicit supervision, making it particularly useful in deployment scenarios common in medical image analysis where distributions shift and labeled samples are unavailable.

\section{Experimental setup}

\addtolength{\tabcolsep}{-0.2pt}
\begin{table*}[t]
\centering
\caption{
\emph{Comparison of various \peft methods for \essa.} 
}
\small
\begin{tabular}{@{} l c ccc ccc ccc ccc @{}}
\toprule
 &  &
\multicolumn{2}{c}{APTOS} && 
\multicolumn{2}{c}{DDSM} && 
\multicolumn{2}{c}{Colorectal} && 
\multicolumn{2}{c}{Average}
\\
\cmidrule(lr){3-4} 
\cmidrule(lr){6-7} 
\cmidrule(lr){9-10} 
\cmidrule(lr){12-13}
Method & Foundation model &
\knn & \sa && 
\knn & \sa && 
\knn & \sa && 
\knn & \sa
\\
\midrule
Standard transfer learning & {ViT-S} + \dinovtwo &
77.2 & 88.9 && 
85.9 & 92.8 && 
87.0 & 96.0 && 
83.4 & 92.6
\\
\full parameter \ssa &  {ViT-S} + \dinovtwo &
80.6 & 89.9 && 
88.3 & 94.1 && 
92.6 & \textbf{96.8} && 
87.2 & 93.6
\\
\essa with \vpt \cite{jia2022visual} & {ViT-S} + \dinovtwo &
69.5 & 89.6 &&
88.6 & 93.7 &&
89.2 & 96.4 && 
82.4 & 93.2
\\
\essa with \bias \cite{zaken2021bitfit} & {ViT-S} + \dinovtwo &
81.1 & \textbf{90.2} &&
89.3 & 94.3 &&
92.0 & \textbf{96.8} && 
87.5 & 93.8
\\
\essa with \lora \cite{hu2021lora} & {ViT-S} + \dinovtwo &
81.7 & 89.4 &&
87.9 & 93.5 &&
92.8 & 96.2 &&
87.5 & 93.0
\\
\essa with \apla \cite{sorkhei2025apla} &  {ViT-S} + \dinovtwo &
\textbf{84.0} & \textbf{90.2} &&
\textbf{91.0} & \textbf{94.9} &&
\textbf{93.6} & \textbf{96.8} &&
\textbf{89.5} & \textbf{94.0}
\\
\midrule
Standard transfer learning &  {ViT-B} + \dinovtwo &
79.0 & 90.2 &&
86.6 & 95.1 && 
88.8 & 97.8 && 
84.8 & 94.4
\\
\full parameter \ssa &  {ViT-B} + \dinovtwo &
81.3 & 90.4 && 
88.7 & 95.2 && 
93.4 & 98.0 && 
87.8 & 94.5
\\
\essa with \vpt \cite{jia2022visual} & {ViT-B} + \dinovtwo &
77.4 & 90.5 &&
88.7 & 95.3 &&
92.0 & \textbf{98.2} &&
86.0 & 94.7
\\
\essa with \bias \cite{zaken2021bitfit} & {ViT-B} + \dinovtwo &
82.3 & 90.5 &&
89.7 & 95.3 &&
94.0 & 98.0 &&
88.7 & 94.6
\\
\essa with \lora \cite{hu2021lora} & {ViT-B} + \dinovtwo &
82.1 & 90.5 &&
89.5 & 95.0 &&
93.4 & 98.0 &&
88.3 & 94.5
\\
\essa with \apla \cite{sorkhei2025apla} &  {ViT-B} + \dinovtwo &
\textbf{84.8} & \textbf{90.8} && 
\textbf{91.9} & \textbf{95.4} &&
\textbf{94.6} & \textbf{98.2} &&
\textbf{90.4} & \textbf{94.8}
\\
\midrule
Standard transfer learning & {ViT-L} + \dinovtwo &
81.2 & 90.5 && 
88.2 & 96.3 && 
89.6 & 98.0 &&
86.3 & 94.9
\\
\full parameter \ssa &  {ViT-L} + \dinovtwo &
83.7 & 90.9 &&
89.3 & 96.5 &&
94.2 & 98.0 &&
89.1 & 95.1
\\
\essa with \vpt \cite{jia2022visual} &  {ViT-L} + \dinovtwo &
78.6 & 91.4 &&
89.2 & 96.2 &&
94.6 & 98.0 &&
87.5 & 95.2
\\
\essa with \bias \cite{zaken2021bitfit} &  {ViT-L} + \dinovtwo &
85.2 & \textbf{91.9} &&
90.1 & 96.7 &&
95.0 & 97.8 &&
90.1 & 95.5
\\
\essa with \lora \cite{hu2021lora} &  {ViT-L} + \dinovtwo &
82.2 & 90.6 &&
89.8 & 96.6 &&
\textbf{96.6} & 98.2 && 
89.5 & 95.1
\\
\essa with \apla \cite{sorkhei2025apla} &  {ViT-L} + \dinovtwo &
\textbf{86.9} & \textbf{91.9} &&
\textbf{92.7} & \textbf{97.0} &&
96.2 & \textbf{98.4} &&
\textbf{91.9} & \textbf{95.8}
\\

\midrule
Standard transfer learning & {ViT-L}  + \imagenettwentyonek &
80.9 & 87.6 && 
86.8 & 93.3 && 
91.8 & 96.4 && 
86.5 & 92.4
\\
\full parameter \ssa &   {ViT-L} + \imagenettwentyonek &
83.1 & 89.0 && 
88.9 & 93.7 && 
95.2 & 97.8 && 
89.1 & 93.5
\\
\essa with \vpt \cite{jia2022visual} & {ViT-L} + \imagenettwentyonek &
79.8 & 89.2 && 
89.4 & 94.4 && 
95.2 & 97.8 && 
88.1 & 93.8
\\
\essa with \bias \cite{zaken2021bitfit} &  {ViT-L} + \imagenettwentyonek &
80.1 & 89.6 && 
90.4 & 94.4 && 
94.4 & 97.6 && 
88.3 & 93.9
\\
\essa with \lora \cite{hu2021lora} & {ViT-L} + \imagenettwentyonek &
79.6 & 88.8 && 
88.2 & 93.4 && 
95.2 & 96.4 && 
87.7 & 92.9
\\
\essa with \apla \cite{sorkhei2025apla} &  {ViT-L} + \imagenettwentyonek &
\textbf{83.2} & \textbf{89.8} && 
\textbf{90.5} & \textbf{94.6} && 
\textbf{96.6} & \textbf{98.0} && 
\textbf{90.1} & \textbf{94.1}
\\
\midrule
Standard transfer learning & {ViT-L} + \dinovtwo &
- & 90.5 && 
- & 96.3 && 
- & 98.0 &&
- & 94.9
\\
\essa with \vpt \cite{jia2022visual}  & {ViT-L} + \dinovtwo &   
- & 91.1$^*$ && 
- & \textbf{97.9}$^*$ && 
- & {98.2}$^*$ && 
- & {95.7}$^*$
\\
\essa with \bitfit \cite{zaken2021bitfit} & {ViT-L} + \dinovtwo & 
- & 90.5$^*$ && 
- & {97.0}$^*$ && 
- & {97.8}$^*$ && 
- & 95.1$^*$ 
\\
\essa with \lora \cite{hu2021lora} & {ViT-L} + \dinovtwo  & 
- & {91.2}$^*$ && 
- & {97.4}$^*$ &&
- & {98.2}$^*$ &&
- & {95.6}$^*$ 
\\
\essa with \apla \cite{sorkhei2025apla} & {ViT-L} + \dinovtwo  & 
- & \textbf{92.0}$^*$ && 
- & {97.1}$^*$ && 
- & \textbf{98.6}$^*$ && 
- & \textbf{95.9}$^*$ 
\\ 
\bottomrule
\multicolumn{13}{l}{$^*$ indicates supervised adaptation (\sa) is also performed with \peft.}
\end{tabular}
\label{tab:ssl_main}
\end{table*}
\addtolength{\tabcolsep}{-0.2pt}

\subsection{Datasets}
We experiment with \NumDatasets medical image analysis datasets.
For classification tasks we use APTOS2019 \cite{aptos2019-blindness-detection}, DDSM \cite{lee2017curated}, and Colorectal \cite{kather2016multi}.
Additionally, we evaluate efficient adaptation methods for TTT \cite{sun2020test} by assessing models trained on APTOS2019 and adapted at test time to IDRiD \cite{porwal2018indian}, 
which presents a domain shift in diabetic retinopathy classification.
For semantic segmentation, we use ISIC2018 \cite{codella2019skin,tschandl2018ham10000}.
We use the provided standard train/val/test splits and performance metrics when available, otherwise following \cite{sorkhei2025apla,matsoukas2022makes}.
We utilize the appropriate evaluation metric for each dataset: 
Quadratic Cohen's Kappa for APTOS2019 and IDRiD, Accuracy for Colorectal, ROC-AUC for DDSM, and Intersection over Union (IoU) for ISIC2018.
In all cases, the same data is available for \essa, \ssa, and \sa stages.

\subsection{Implementation details}
For self-supervision we use \dinovtwo \cite{oquab2023dinov2} using a \vit \cite{dosovitskiy2020image} backbone.
For semantic segmentation we use the SETR-PUP framework \cite{zheng2021rethinking} with a \vitl encoder and convolutional decoder.
The \peft methods we consider are
\lora \cite{hu2021lora} which injects low-rank trainable matrices into self-attention layers, \vpt \cite{jia2022visual} which modifies input token embeddings using learnable prompts, 
\bitfit \cite{zaken2021bitfit} which fine-tunes only bias terms, and 
\apla \cite{sorkhei2025apla} which selectively fine-tunes a small fraction of projection layer weights in self-attention blocks.
For \essa, models are trained for 300 epochs using AdamW \cite{adamw}, with hyperparameters as recommended by \cite{oquab2023dinov2}. 
For \sa, models are trained for 100 epochs, utilizing a cosine decay schedule with a 10-epoch warm-up.
Hyperparameters are selected via grid search on the validation set. 
\essa is conducted on 4 A100 GPUs in a multi-GPU training setup with an effective batch size of 256, while \sa is performed on a single A100 GPU with a batch size of 64.

\subsection{Evaluation protocols}
To evaluate \essa we follow two widely used evaluation protocols \cite{caron2021emerging,oquab2023dinov2}:
\textit{1)} We use a weighted nearest neighbor classifier (\knn) with the adapted backbone as the feature extractor to evaluate representation quality.
\textit{2)} A classification head is appended to the adapted backbone, and the entire network undergoes \sa.
This evaluation framework remains consistent across full-network training and all \peft methods. 
For test-time training, the classification head trained with \sa is frozen and used for prediction on test data after applying \essa to adapt the feature extractor to the test data.

\section{Results}

Our main results in Table \ref{tab:ssl_main} demonstrate that efficient adaptation methods offer, on average, an improvement in downstream performance over full-parameter SSA, but the gains are not consistent across \peft methods.
Among the \peft methods, \apla demonstrates consistent improvements over standard transfer learning and full-parameter \ssa across different tasks, architectures, and evaluation protocols.
Representation quality, as measured by \knn paints a similar picture, where most \essa methods (aside from \vpt) learn better representations than full-parameter \ssa.
We observe a trend of diminishing returns  from full-parameter \ssa as the foundation model grows, while \essa methods continue to improve adaptation performance.
For sufficiently large models, \essa is universally better than \ssa. 
Interestingly, all \peft methods benefit from larger foundation models. These results demonstrate that \peft can be effectively applied to \ssa to enhance self-supervised learning, revealing a previously unexplored potential of these methods.
This trend persists when using a ViT-L pretrained on \imagenettwentyonek.

\subsection{Integrating \peft across both SSA and SA}
\begin{figure}[t]
    \centering
    \includegraphics[width=\linewidth]{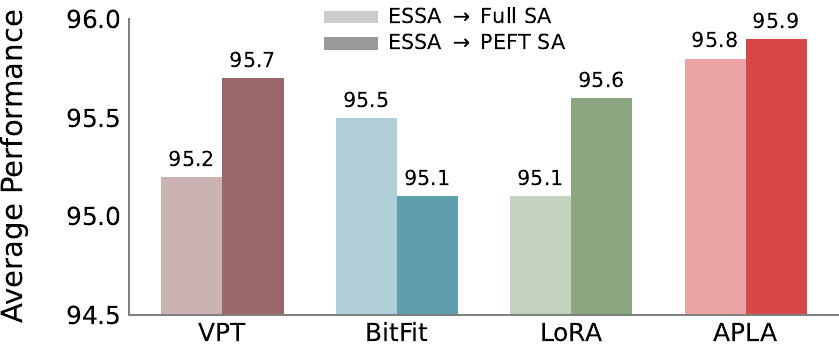}
    \caption{
    \emph{\essa followed by efficient \sa.}
    }
    \label{fig:ssa_sa_both_eff}
\end{figure}

We investigate the application of \peft methods throughout the entire training pipeline—performing \ssa with \peft and subsequently continuing with \peft during \sa.
When we apply the \peft methods to both the \ssa and \sa steps, we see additional gains in downstream performance for all methods aside from \bitfit.
(final block of Table \ref{tab:ssl_main} and Figure \ref{fig:ssa_sa_both_eff}).
These findings suggest that integrating \peft across the whole two-stage training pipeline yields the best results.

\begin{figure}[t]
    \centering
    \includegraphics[width=\linewidth]{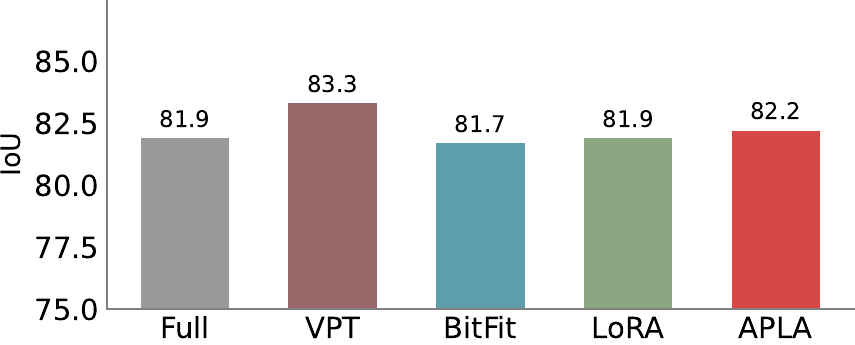}
    \caption{
    \emph{\essa for semantic segmentation.}
    }
    \label{fig:seg}
\end{figure}

\subsection{\essa for semantic segmentation}
In addition to medical image classification tasks, we also evaluate the effectiveness of \essa in the context of medical image segmentation.
Semantic segmentation results on ISIC2018  are given in Figure \ref{fig:seg}, where \essa is performed followed by full-parameter SA. 
Notably, both \vpt and \apla outperform full-parameter SSA, while \lora achieves comparable performance, demonstrating the competitiveness of \peft methods in this setting.

\subsection{\peft for test-time training (TTT)}
\begin{figure}[t]
    \centering
    \includegraphics[width=\linewidth]{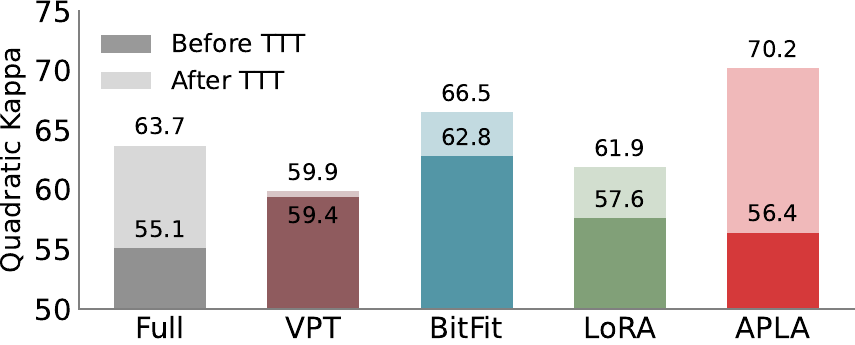}
    \caption{
    \emph{TTT on OOD data using \peft.}  
    }
    \label{fig:ttt}
\end{figure}

We also investigate the effectiveness of \peft for test-time training (TTT), where the model is further adapted to distribution shifts at inference time using self-supervision.
In Figure \ref{fig:ttt}, we apply a \peft method to our entire \ssa/\sa pipeline on a source domain (APTOS2019), then adapt the feature extractor to a new domain at test time \cite{sun2020test} (IDRiD) using the same \peft method.
Results show the difference between the source-adapted model and the model adapted to the test data using TTT.
Among the \peft methods, \bitfit and \apla demonstrate superior performance compared to full-parameter tuning after TTT.

\subsection{Computational efficiency}
\begin{figure}[t]
    \centering
    \includegraphics[width=\linewidth]{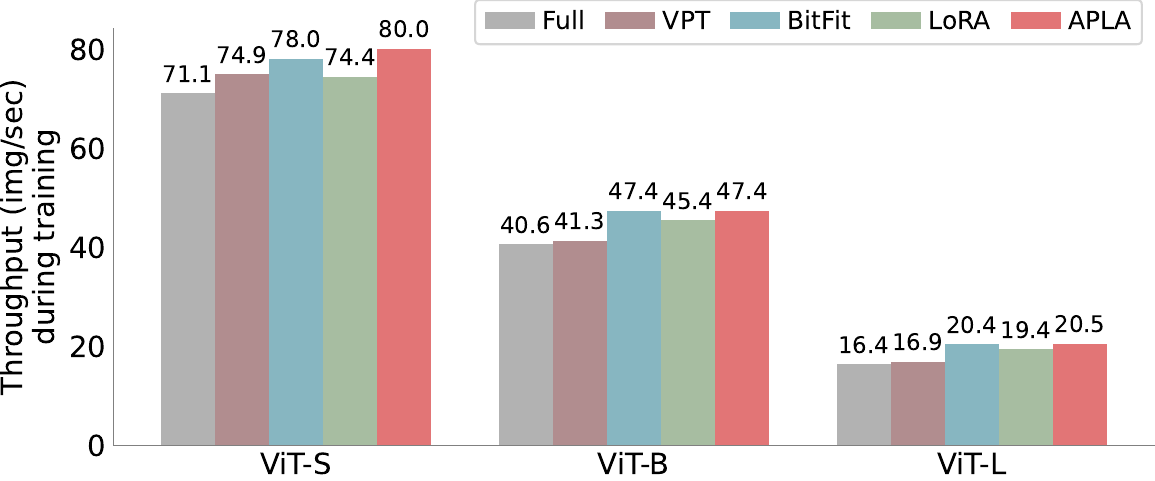}
    \caption{
    \emph{\ssa/\essa training throughput.}
    }
    \label{fig:throughput}
\end{figure}

We report GPU memory usage in Figure \ref{fig:fig1}e and training throughput in Figure \ref{fig:throughput}. 
All \peft methods demonstrate substantial efficiency improvements over full-parameter tuning, with particularly notable gains observed for \apla — achieving \MemReduction savings in GPU memory consumption and \ThroughputIncrease training speed-up.
These results highlight the effectiveness of \peft for reducing the computational burden of \ssa.

\section{Discussion}

\noindent \textit{\essa challenges the assumption that full-parameter \ssa is necessary.}
\peft methods were originally designed for standard transfer learning (\sa), 
leading to the natural question of whether they are as effective for \ssa.
One key concern is that optimizing only a small subset of parameters may not be enough during \ssa for adapting the rich foundational representations to medical  tasks.
Our results show that \ssa provides diminishing returns for large foundation models, with only marginal improvements over standard supervised transfer learning (\sa). 
This suggests that the primary limitation of \ssa for large models is not just the computational cost, but also limited adaptation gains.
However, applying {\peft}s in \essa fundamentally shifts this paradigm: once a foundation model is sufficiently large, \peft methods outperform full-parameter \ssa, making \essa the better strategy. Instead of a costly adaptation step with marginal benefits, \essa makes \ssa both feasible and effective.

\vspace{2mm}
\noindent \textit{\peft enhances domain adaptation at all stages.}
Beyond \ssa, our results demonstrate that \peft provides the greatest gains when applied to \sa and TTT as well as \ssa. 
This result was not trivial at the outset--yet our findings suggest that \peft is not just an efficiency trade-off, but an actively better approach than full-parameter tuning for domain adaptation. 
By enforcing structured adaptation, \peft methods facilitate better feature reuse between \ssa and \sa, while also improving on-the-fly adaptation in test-time training (TTT). 
From a practical perspective, this means that \peft enables performant and computationally efficient medical AI, allowing large foundation models to adapt dynamically, even in resource-constrained deployment settings  where labeled data is unavailable.

\vspace{2mm}
\noindent \textit{\apla is the best \essa method across all scenarios.}
While all \peft methods improve efficiency, we found that \apla is the only approach that consistently brings performance benefits, outperforming full-parameter \ssa across all tasks and architectures. 
By the simple virtue of limiting the number of tunable parameters, \apla is incentivized to better preserve the rich representations of the pretrained foundation models.
While other \peft methods may also be benefiting from this same implicit regularization effect, 
\apla is unique in that it  forces the model to recompose existing features within the critical attention projection layers \cite{sorkhei2025apla}. 
This implicit regularization may be particularly beneficial in medical imaging, where data is often scarce, and models risk memorizing low-variance distributions dominated by background pixels.

\vspace{2mm}
\noindent \textit{Implications for Medical AI.}
The success of \essa highlights a major shift in how adaptation should be approached for large-scale medical AI. 
Our results suggest that targeted, efficient adaptation can match or exceed traditional transfer learning and even the more powerful \ssa approaches at a fraction of the cost. 
This enables development of specialized medical AI models in resource-constrained settings.
Furthermore, the reduced computational requirements lower barriers to entry for medical institutions seeking to implement AI-driven solutions tailored to their specific patient populations and clinical workflows.

\vspace{2mm}
\noindent \textit{Limitations and future work.}
While our findings establish \essa as an effective paradigm, there are still open questions. 
Our study primarily focused on \dinovtwo as the self-supervised learning (SSL) objective, as initial experiments showed it to be superior to other SSL methods. 
However, we did not systematically investigate whether other contrastive or masked image modeling approaches might yield different adaptation behaviors with \peft. 
Future work should explore whether certain SSL paradigms are better suited for \essa and whether additional architectural modifications can further improve adaptation efficiency.

\section{Conclusion}
In this work we investigate the effectiveness of \essa for medical image analysis.
We show that applying \peft methods improve the quality of adapted representations and boost downstream performance while reducing computational costs.
Furthermore, we demonstrate that applying \peft across the entire two-stage adaptation pipeline yields the best results.
By demonstrating the effectiveness and efficiency of \peft methods for self-supervised adaptation, we aim to make foundation models more accessible to the wider medical image analysis community.

\paragraph{\textbf{Acknowledgments.}}
This work was supported by the Wallenberg AI, Autonomous Systems and Software Program (WASP).
We acknowledge the Berzelius computational resources provided by the Knut and Alice Wallenberg Foundation at the National Supercomputer Centre and the the computational resources provided by the National Academic Infrastructure for Supercomputing in Sweden (NAISS), partially funded by the Swedish Research Council through grant agreement no. 2022-06725.

{
    \small
    \bibliographystyle{ieeenat_fullname}
    \bibliography{References}
}

\end{document}